\documentclass[letterpaper]{article} 
\usepackage{aaai25}  
\usepackage{times}  
\usepackage{helvet}  
\usepackage{courier}  
\usepackage[hyphens]{url}  
\usepackage{graphicx} 
\usepackage{natbib}  
\usepackage{caption} 
\usepackage{algorithm}
\usepackage{algorithmic}
\usepackage{amsmath}

\usepackage{multirow} 
\usepackage{amssymb} 
\usepackage{booktabs} 
\usepackage{textcomp} 

\usepackage{newfloat}
\usepackage{listings}
\DeclareCaptionStyle{ruled}{labelfont=normalfont,labelsep=colon,strut=off} 
\floatstyle{ruled}
\newfloat{listing}{tb}{lst}{}
\floatname{listing}{Listing}
\title{CTA-Net: A CNN-Transformer Aggregation Network for Improving Multi-Scale Feature Extraction}
\author{
    Chunlei Meng\textsuperscript{\rm 1},
    Jiacheng Yang\textsuperscript{\rm 2},
    Wei Lin\textsuperscript{\rm 1},
    Bowen Liu\textsuperscript{\rm 1},
    Hongda Zhang\textsuperscript{\rm 1},\\
    Zhongxue Gan\textsuperscript{\rm 1}\equalcontrib,
    Chun Ouyang\textsuperscript{\rm 1}\equalcontrib
}
\affiliations{
    \textsuperscript{\rm 1}Fudan University Academy for Engineering and 
Technology\\
    \textsuperscript{\rm 2}Jihua Laboratory\\

    \{clmeng23, 22210860047, bwliu22\}@m.fudan.edu.cn, \{zhanghongda, ganzhongxue, {oy\_c}\}@fudan.edu.cn \\yangjch@jihualab.ac.cn
}

\usepackage{bibentry}

\begin{document}
\maketitle

\begin{abstract}
Convolutional neural networks (CNNs) and vision transformers (ViTs) have become essential in computer vision for local and global feature extraction. However, aggregating these architectures in existing methods often results in inefficiencies. To address this, the CNN-Transformer Aggregation Network (CTA-Net) was developed. CTA-Net combines CNNs and ViTs, with transformers capturing long-range dependencies and CNNs extracting localized features. This integration enables efficient processing of detailed local and broader contextual information. CTA-Net introduces the Light Weight Multi-Scale Feature Fusion Multi-Head Self-Attention (LMF-MHSA) module for effective multi-scale feature integration with reduced parameters. Additionally, the Reverse Reconstruction CNN-Variants (RRCV) module enhances the embedding of CNNs within the transformer architecture. Extensive experiments on small-scale datasets with fewer than 100,000 samples show that CTA-Net achieves superior performance (TOP-1 Acc 86.76\%), fewer parameters (20.32M), and greater efficiency (FLOPs 2.83B), making it a highly efficient and lightweight solution for visual tasks on small-scale datasets (fewer than 100,000).



\end{abstract}

%

\begin{figure}[htbp]
\centering
\includegraphics[width=0.85\columnwidth]{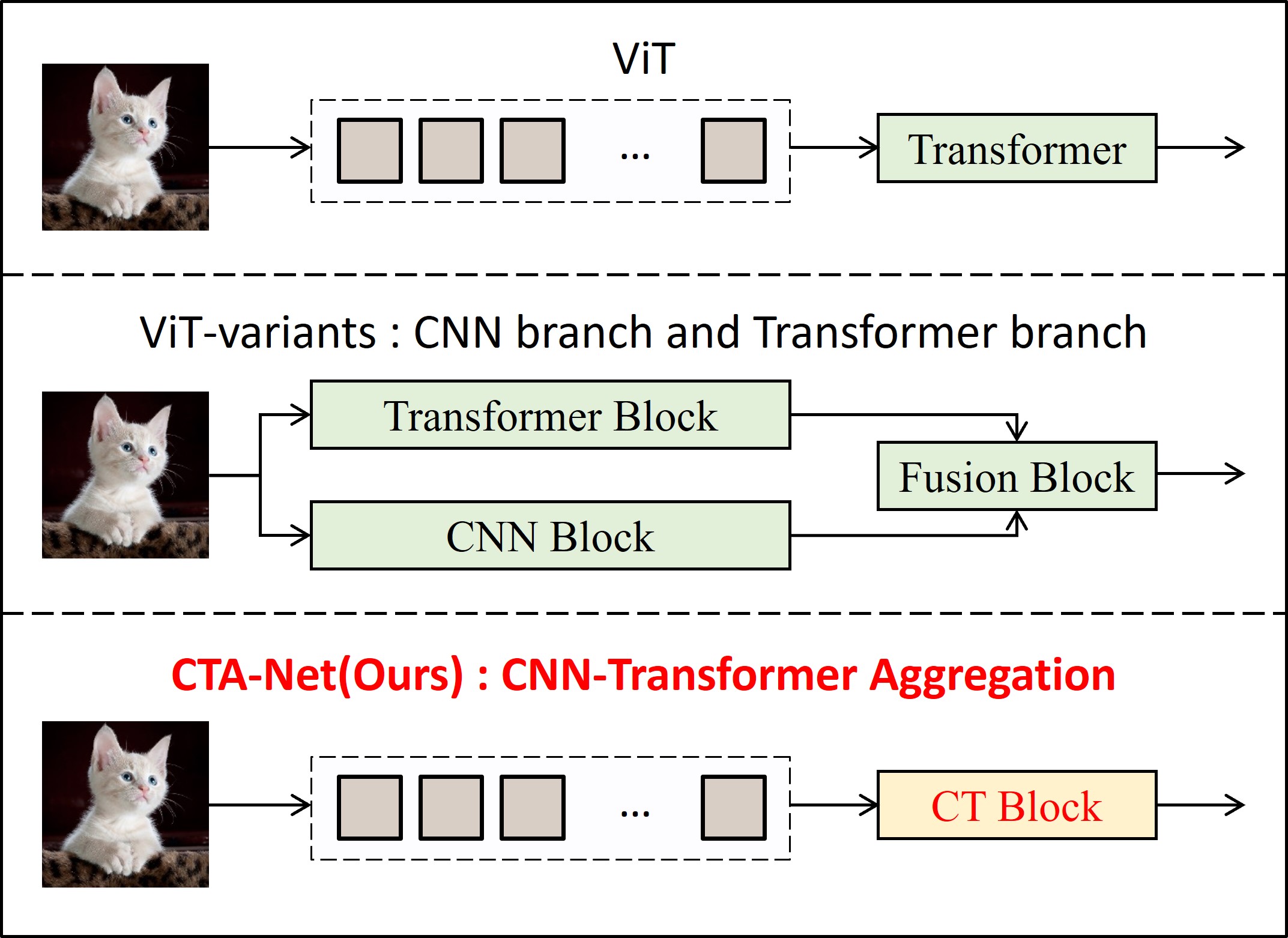}
\caption{\textbf{Top}: ViT use pure Transformer block. \textbf{Middle}: The state-of-the-art ViT-Variants use CNN branch and Transformer branch. \textbf{Bottom}: CTA-Net use CNN-Transformer Aggregation Network, aggregate CNN into Transformer to take full advantage of the advantages of both.}
\label{fig1}
\end{figure}

\section{Introduction}

Convolutional neural networks (CNNs) have been at the forefront of advancements in computer vision due to their powerful ability to extract detailed, discriminative features \cite{he:2016deep,7_0813_ouyang2021peripheral,8_0813_ouyang2021inter}. By employing convolutional layers, CNNs efficiently capture local spatial hierarchies, contributing to state-of-the-art performance in various image classification tasks. Despite their effectiveness in local feature extraction, the inherent limitation of CNNs lies in the constrained receptive field of small convolutional kernels, which can impede the capture of global contextual information. To address this limitation, researchers often incorporate additional mechanisms or layers to capture a more comprehensive visual context \cite{13:ngo2024learning,36:guo2022cmt}.

Self-attention-based transformers, such as the Vision Transformer (ViT) \cite{dosovitskiy2020image}, have emerged as a compelling alternative to CNNs, primarily because of their capability to model long-range dependencies. ViT segments an image into patches, transforming them into a sequence of tokens akin to word tokens in natural language processing (NLP). These patches, supplemented by positional embeddings, are fed into stacked transformer blocks to model global relations and extract classification features. The self-attention mechanism, a core component of ViT, enables the network to capture extensive spatial dependencies within images \cite{6:yoo2023enriched}.

However, existing transformer-based models \cite{liang2021swinir,chen2021pre,6:yoo2023enriched} face challenges in leveraging local and multi-scale features, which are crucial for many visual tasks \cite{7:guo2020closed,8:mei2020image}. Two primary concerns arise when building transformer-based architectures: firstly, although ViT effectively captures long-range dependencies between image patches \cite{36:guo2022cmt}, it may neglect the spatial local information within each patch—an area where CNNs excel \cite{9:d2021convit,10:wu2021cvt}. Secondly, the uniform size of tokens in ViT restricts the model's ability to exploit multi-scale relationships among tokens, which is particularly beneficial for various downstream tasks \cite{8:mei2020image,11:shocher2018zero}.

Both ViTs and CNNs architectures bring distinct advantages to the table. When integrated effectively, they can leverage their respective strengths to improve model performance \cite{13:ngo2024learning}. While ViT has shown robustness in capturing global representations, particularly with large datasets, it is prone to overfitting on small-scale datasets (fewer than 100,000) due to its reliance on multi-layer perception (MLP) layers \cite{14:popescu2009multilayer}. Conversely, CNNs are adept at capturing local representations and exhibit robust performance on small-scale datasets but may not scale as efficiently to larger datasets.

This work proposes a new method that integrates the complementary strengths of CNNs and ViTs without increasing unnecessary computation. As illustrated in Figure \ref{fig1}, the proposed CNN-Transformer Aggregation Network (CTA-Net) enhances the capabilities of ViTs by incorporating CNNs as an integral component, compensating for the limitations of pure transformer-based models.

In summary, The main contributions of this paper are as follows:

\begin{itemize}
\item The Reverse Reconstruction CNN-Variants (RRCV) module is seamlessly integrated into transformer architectures, combining the local feature extraction capabilities of CNNs with the global contextual understanding of ViTs.

\item The Lightweight Multi-Scale Feature Fusion Multi-Head Self-Attention (LMF-MHSA) module efficiently leverages multi-scale features while maintaining a reduced parameter count, enhancing model efficiency and performance, particularly in resource-constrained environments.
\end{itemize}

\section{Related Works}

\subsection{CNN and Transformer Aggregation Network}

The aggregation of CNNs and ViTs has become a key focus in contemporary research \cite{10_0813_vasu2023fastvit}, as researchers explore the synergistic combination of CNNs' local feature extraction capabilities with ViTs' global contextual understanding \cite{20:lu2024sbcformer,38:vasu2023fastvit}. Various methods have been developed to blend these strengths, such as the Swin Transformer \cite{24:swin-t}, which uses windowed attention mechanisms for implicit local and global feature integration. Other approaches incorporate explicit fusion structures to facilitate the exchange of information between tokens or patches, creating a more unified feature representation \cite{10:wu2021cvt,36:guo2022cmt,9:d2021convit}.

In typical aggregation architectures, CNNs and Transformers are organized into dual branches that independently learn before integration. For example, Dual-ViT \cite{44:yao2023dual} uses two distinct pathways to capture both global and local information. ECT \cite{6:yoo2023enriched} introduces a Fusion Block to bidirectionally connect intermediate features between CNN and Transformer branches, enhancing the strengths of each. SCT-Net \cite{37:xu2024sctnet} proposes a dual-branch architecture where CNN and Transformer features are aligned to encode rich semantic information and spatial details, which the Transformer utilizes during inference. Crossformer++ \cite{31:crossformer++} extends this concept with a pyramid structure inspired by CNNs to hierarchically expand channel capacity while reducing spatial resolution.

Despite these advances, such architectures often treat CNNs and Transformers as separate modules that interact superficially, necessitating fusion blocks or similar structures to assist feature integration. This separation can hinder information flow between the two, potentially leading to information loss. Moreover, for small-scale datasets where the features for learning are limited, these fusion architectures may restrict comprehensive feature learning \cite{36:guo2022cmt}. This limitation is particularly problematic in tasks requiring detailed local features and comprehensive global context, such as image classification.

\subsection{Multi-Head Self-Attention Mechanism}

The Multi-Head Self-Attention (MHSA) mechanism is crucial for capturing global dependencies across spatial positions, significantly enhancing Transformer performance in visual tasks \cite{4_0813_yguo2020multi}. However, many MHSA mechanisms rely on single-scale learning processes, which restrict the model's capacity to capture multi-scale features \cite{1_0813_yan2024multi}. This limitation is particularly evident in tasks requiring a nuanced understanding of both global context and local features \cite{5_0813_yhao2021self}. For instance, single-scale MHSA models often fail to exploit the varying levels of granularity in the data, leading to suboptimal feature representation and impairing performance in tasks such as image classification or object detection \cite{2_0813_wang2023ultra,3_0813_hao2021self}.

Recent advancements aim to address these deficiencies by developing multi-scale MHSA models \cite{6_0813_luo2024dual}. Cross-ViT \cite{26:cross-vit} introduces an innovative architecture that encodes and fuses multi-scale features, enhancing the model's ability to leverage various levels of detail from input data. SBCFormer \cite{20:lu2024sbcformer} achieves high accuracy and fast computation on single-board computers by introducing a new attention mechanism.

The LCV model \cite{13:ngo2024learning} addresses domain adaptation challenges by combining CNNs' local feature extraction with ViTs' global context understanding. However, the performance is not ideal when faced with small-scale datasets with limited features.

These complexities underscore the ongoing challenge of designing efficient Transformer architectures that effectively capture multi-scale features without incurring prohibitive computational costs. Addressing this issue remains a critical area of research, particularly for applications involving small-scale datasets where comprehensive feature learning is paramount \cite{9_0813_Gani_2022_BMVC}.

\begin{figure*}[!ht]
\centering
\includegraphics[width=0.90\linewidth]{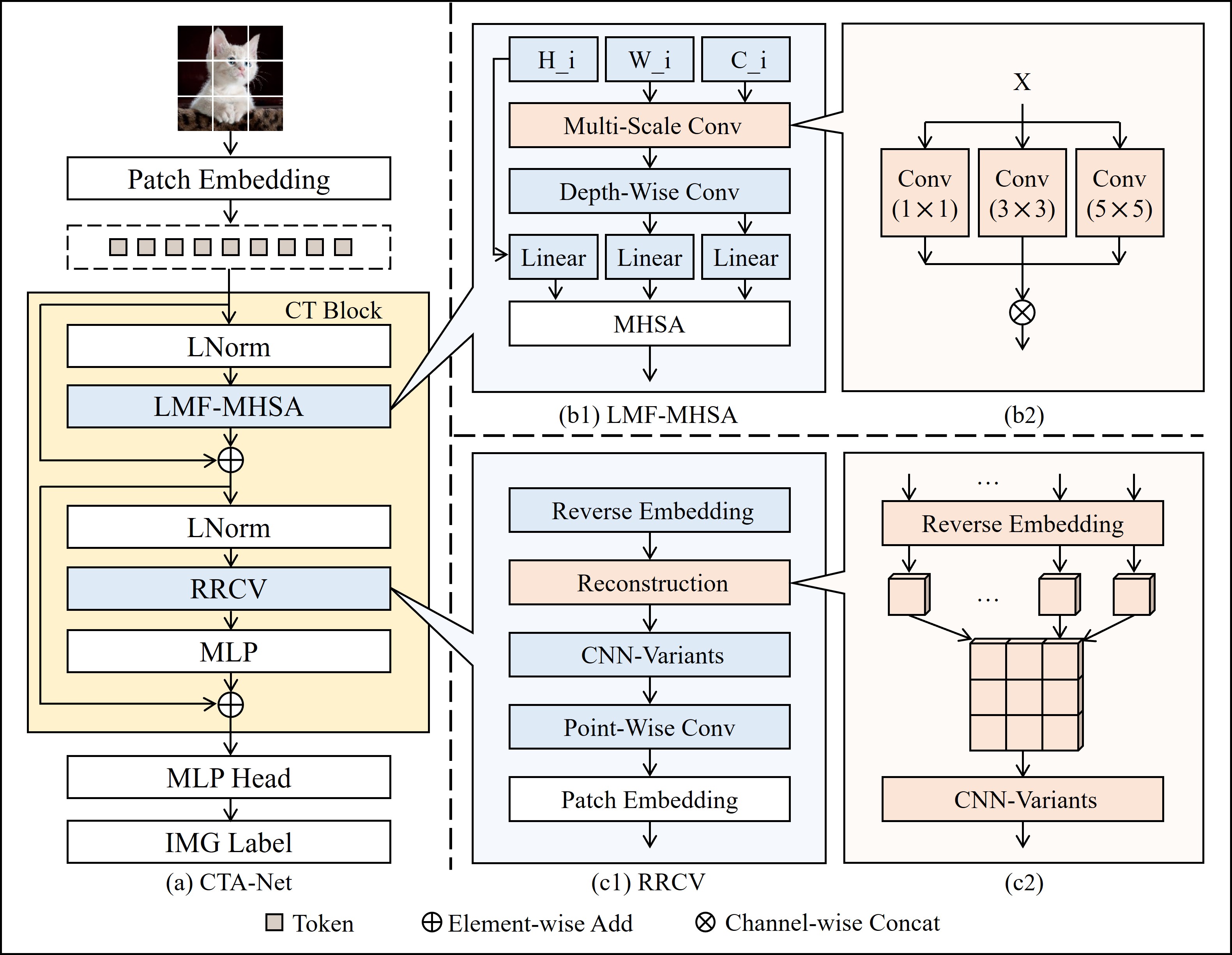}
\caption{\textbf{(a)} illustrates \textbf{the overall architecture of CTA-Net}, highlighting the central CNN-Transformer (CT) Block, which integrates CNNs with transformers for enhanced feature extraction.
\textbf{(b1)} depicts the \textbf{LMF-MHSA module}, showcasing the Lightweight Multi-Scale Feature Fusion Multi-Head Self-Attention mechanism, which efficiently learns multi-scale features while reducing computational complexity. \textbf{(b2)} provides a detailed view of the \textbf{Multi-Scale Conv operation}, demonstrating how different convolution kernel sizes are used to extract multi-scale features from the input. \textbf{(c1)} illustrates the \textbf{RRCV module}, the Reverse Reconstruction CNN-Variants module, designed to embed CNN operations within the Transformer architecture, leveraging the strengths of both CNNs and transformers. \textbf{(c2)} offers a detailed view of the Reconstruction operation process, highlighting how local features extracted by CNNs are seamlessly integrated into the Transformer's global context.}
\label{fig2}
\end{figure*}

\section{Method}
This section provides a concise overview of the proposed CTA-Net network architecture, followed by a detailed introduction to its components. 

\subsection{Overall Architecture}
The intention is to construct an aggregation network that leverages the advantages of both CNNs and Transformers. As illustrated in Figure \ref{fig2}, the CTA-Net is designed to integrate the strengths of CNNs and ViTs. The architecture incorporates two key modules: the RRCV and the LMF-MHSA. These modules ensure a seamless blending of local and global features while maintaining computational efficiency.

In the proposed CTA-Net, the input image is divided into patches, which are transformed into a sequence of tokens. These patches are embedded into a high-dimensional space, similar to the token embedding process in ViTs. Positioned after the initial Layer Normalization (LNorm) module, the LMF-MHSA module replaces the conventional Multi-Head Self-Attention (MHSA) mechanism, efficiently handling multi-scale feature fusion while reducing computational complexity and memory usage. This is achieved by considering different scales of the input tokens, thereby reducing the computational load compared to traditional MHSA. Located after the second LNorm module and before the Multi-Layer Perceptron (MLP) module in the Transformer block, the RRCV module integrates CNN operations into the Transformer. This module enhances local feature extraction through convolution operations and reconstructs these features to blend with the Transformer's global context, ensuring the local details captured by CNNs are effectively utilized within the Transformer architecture. The sequence of tokens then passes through multiple Transformer blocks, each consisting of the LMF-MHSA and RRCV modules, ensuring comprehensive feature extraction at both local and global levels by leveraging the strengths of CNNs and Transformers. Finally, the token representations are fed into a classification head to perform the desired vision task, such as image classification. By fully integrating CNNs and Transformers, CTA-Net effectively captures both local and global features, leading to more comprehensive and accurate feature representation, reduced computational complexity, and improved performance. Extensive experiments on benchmark datasets demonstrate that CTA-Net outperforms existing methods in various vision tasks, providing a robust and practical solution for real-world applications.

\subsection{Reverse Reconstruction CNN-Variants}
CNNs have historically excelled in various computer vision tasks by effectively capturing local features among adjacent pixels. Throughout their evolution, numerous variant architectures have emerged, such as ResNet \cite{he:2016deep} and depth-wise separable convolutions \cite{18:tan2019efficientnet}. These innovations have addressed specific challenges inherent to deep networks, such as mitigating the degradation problem that arises with increasing depth and reducing the excessive parameterization typically associated with traditional convolutional networks.

The RRCV module's integration into CTA-Net follows a multi-step process, as illustrated in Figure \ref{fig2}(c1). Initially, a reverse embedding function $RE(\cdot)$is applied to the Transformer-generated vectors $X_T$, reconstructing them into feature maps $X_{f\_map}$ that align with the input specifications of convolutional neural networks. Subsequently, Point-Wise Convolution ($PConv(\cdot)$) is utilized to effectively reduce data dimensionality and computational complexity. The final step involves using the patch embedding function $E(\cdot)$ to integrate these processed vectors back into the Transformer framework seamlessly, avoiding the need for an intermediate fusion block that could potentially cause information loss. This process is formally expressed as follows:

\begin{equation}
    X_{f\_map} = CNN(RE(X_{T}))
\end{equation}
\begin{equation}
    X_{RRCV} = E(PConv(X_{f\_map}))
\end{equation}

\subsubsection{Reconstruction}
As depicted in Figure \ref{fig2}(c2), the reconstruction process is designed to recover the intermediate results of the Transformer into the original feature maps, preserving the corresponding positional information through position embedding combination. These reconstructed feature maps are then processed by our designed CNN-Variants module. Here, $R^{C \times H \times W}$represents a tensor describing a feature map with dimensions $C,H,W$ and $R^{C \times N \times H_p \times W_p}$represents a patch with dimensions $C,N,H_p,W_p$

\begin{equation}
    R^{C \times H \times W} = Reconstruct(R^{C \times N \times H_p \times W_p}, H, W)
\end{equation}

By avoiding the necessity for a distinct fusion block, the proposed architecture facilitates a seamless blending of CNN and ViT components, ensuring that feature extraction and integration occur without the information loss that can accompany intermediate processing stages. This seamless blending results in a more cohesive and efficient model architecture, effectively harnessing the strengths of both CNN and ViT to achieve superior performance in visual recognition tasks.

\subsubsection{CNN-Variants}

The CNN-Variants module is designed to enhance the ViT's ability to capture local spatial details, which are otherwise limited due to its patch-based approach. By reconstructing ViT vectors into feature maps, the module enables effective local information extraction, subsequently integrating these features with the ViT's global context. To validate the effectiveness of local feature extraction, This paper examines three specific variants of CNNs: standard CNN, residual modules, and depth-wise separable convolution modules. The standard CNN serves as a baseline, illustrating the effectiveness of traditional convolutional methods in extracting local features. Residual modules are selected for their ability to mitigate the vanishing gradient problem in deep networks, thereby enhancing the model's feature learning capacity. Depth-wise separable convolution modules are employed for their efficiency in reducing parameter count while preserving feature extraction accuracy which are critical consideration in resource-constrained environments.

These variants allow for a systematic evaluation of how different convolutional strategies can optimize the integration of local and global features within a Transformer framework.

\subsection{Light Weight Multi-Scale Feature Fusion Multi-Head Self-Attention}

The LMF-MHSA module addresses computational complexity and multi-scale feature extraction challenges in modern computer vision tasks. Traditional MHSA mechanisms are resource-intensive and struggle to capture features across multiple scales, leading to suboptimal object detection. The proposed LMF-MHSA, as shown in Figure \ref{fig2}(b1), significantly reduces computational costs while enhancing feature extraction through a multi-scale fusion mechanism.


\subsubsection{Multi-scale Feature Fusion}

As shown in Figure \ref{fig2}(b2), The multi-scale feature fusion layer is used to extract features at different scales from the input to improve the model's sensitivity to various scale features. Given an input feature map \( X \), multi-scale features was extracted by using different convolution kernel sizes:
\begin{equation}
X_s = \text{Concat}(\text{Conv}(X_1), \text{Conv}(X_3), \text{Conv}(X_5))
\end{equation}
where \( X_1 \), \( X_3 \), and \( X_5 \) represent the feature maps processed by \( 1 \times 1 \), \( 3 \times 3 \), and \( 5 \times 5 \) convolution kernels, respectively.

\subsubsection{Lightweight Multi-head Self-attention Mechanism}

The LMF-MHSA mechanism introduces several innovative approaches to enhance computational efficiency while preserving model performance:

\textbf{Depthwise Separable Convolution.} This operation decomposes standard convolutions into depthwise and pointwise steps, significantly reducing the number of parameters and the computational load. A conventional convolutional layer with parameters \(M \times N \times D \times D\) is transformed into a more efficient structure with \(M \times D \times D + M \times N\) parameters.

\textbf{Query, Key, and Value Linear Projections.} To optimize resource usage, \(1 \times 1\) convolutions replace traditional matrix multiplications for transforming the Query, Key, and Value matrices, ensuring data integrity with reduced computational cost.

\textbf{Attention Computation and Projection.} The core attention mechanism is defined by:
\[
\text{Attention}(Q, K, V) = \text{Softmax}\left(\frac{QK^T}{\sqrt{d_k}}\right) V
\]
where \(d_k\) represents the dimensionality of the key. Additional linear projections are applied:
\[
\begin{aligned}
K' &= \text{Linear}(K)\\
V' &= \text{Linear}(V)
\end{aligned}
\]
This approach focuses computational efforts on the most pertinent features, balancing precision and efficiency.

\textbf{Output Features and Efficiency.} The LMF-MHSA output is calculated by integrating attention weights with the transformed value vectors:
\[
\text{LMF-MHSA}(X) = \text{Attention}(Q, K', V')
\]

Through a structured process ranging from initial convolutional refinement to optimized attention calculation the LMF-MHSA mechanism effectively captures both local and global features. This makes it particularly suitable for tasks involving small-scale datasets (fewer than 100,000) and constrained computational resources.

\begin{table*}[ht]

\centering

\begin{tabular}{c c c c c c c c c}
\toprule[2pt] 

{\multirow{2}*{\textbf{Type}}} &
{\multirow{2}*{\textbf{Model}}} &
{\multirow{2}*{\textbf{Reference}}} &
{\textbf{APTOS}} &
{\textbf{RFMiD}} & 
{\textbf{CIFAR}} &
{\textbf{CIFAR}} & 
{\textbf{FLOPs}} &
{\textbf{Params}}

\\

 & & &2019&2020&10 &100 &(B)&(M) \\
\hline\hline

\multirow{3}*{CNN-Variants} 
&ResNet34 &CVPR 2016& 73.22\% &  72.81\% &60.74\% &33.75\% & 7.34 & 21.29  \\
&ResNet101 &ICLR 2016& 70.77\% &  80\% & 68.22\%& 40.10\%&  7.80&  44.60\\
&EfficientNet&IMCL 2019& 75.68\% &  77.97\% &65.91\% &36.58\% &19  & 43 \\
\hline
\multirow{4}*{ViT-Variants} &ViT/16 &ICLR 2020 & 64.21\% & 80.62\% & 67.42\%& 41.41\% &11.28 &57.96 \\

&SwinT &CVPR 2022& 68.3\% & 81.1\% &79.3\% &\textbf{63.59\%} &8.70&49.56 \\
&MIL-VT &MICCAI 2021&61.95\%  & 72.8\% &49\% &34.5\% &12.52 &21.72 \\ 

\hline
\multirow{7}*{ViT-Aggregation}&Dual-ViT &TPAMI 2023& 70.77\% & 80.63\% &\textbf{86.94\%} & -  & 5.40 &27.59  \\
&Cross ViT &ICCV 2021& 69.13\%  &80.01\% &75\% & 49.94\%& 5.60 &26.79 \\ 
&CMT &CVPR 2022 & 76.77\% & 80.61\% &-&- &4&25.10\\ 
&SBCF&WACV 2024 & 71.3\% & 80.62\% &- &- &2.91 &20.68\\
&fastViT &ICCV 2023& 74.72\% & 79.37\% &84.72\% & 59.2\% &22 & 46\\
&LCV &CVPR 2024& 72.95\% & 80.66\% & -& -&16.86 &86.98\\
&CrossF++ &TPAMI 2023& 47\% & 79.38\% & 80.49\%& - & 4.08&23.33\\

&\textbf{CTA-Net} &\textbf{Ours}&\textbf{ 76.9\% } &  \textbf{ 82.03\%} & \textbf{86.76\%}&\textbf{59.43\%}  & \textbf{2.83}&\textbf{20.32}\\
\toprule[2pt]

\end{tabular}

\caption{\textbf{Comparison with State-Of-The-Art Methods on small-scale datasets} (fewer than 100,000) (Metric:TOP-1 Acc)}
\label{table_1}
\end{table*}

\section{Experiments}
This section outlines the comprehensive experiments conducted to assess the effectiveness of the proposed CTA-Net and its individual components. Comparative evaluations were performed on benchmark datasets against SOTA approaches. The datasets and implementation details are presented first, followed by a series of ablation studies to validate the performance of individual modules. Finally, comparative experiments illustrate the superiority of CTA-Net over existing SOTA methods.

\subsection{Datasets and Implementation Details}

\subsubsection{Datasets}

ViT and its variants perform well when pre-trained on large-scale datasets, but tend to perform poorly on small-scale datasets (fewer than 100,000) without such pre-training. In contrast, CNN performs well on small-scale datasets, but ViT tends to perform poorly when dealing with small-scale datasets. To verify that CTA-Net fully exploits the advantages of both architectures, the proposed CTA-Net is evaluated on four small-scale datasets.. The four open-source small-scale datasets include CIFAR-10, CIFAR-100 \cite{16:cifar}, APTOS 2019 Blindness Detection (APTOS2019) \cite{40:APTOS2019}, and 2020 Retinal Fundus Multi-Disease Image Dataset (RFMiD2020) \cite{41:rfmid}. The dataset details are shown in Appendix A. To enhance the diversity of training data, a series of data augmentation techniques are applied, including random cropping, rotation, horizontal flipping, and color jittering.

\subsubsection{Implementation Details}

The experiments were conducted to evaluate the feature self-learning capabilities of CTA-Net and the integration of CNN and Transformer components without utilizing pre-trained weights. Model performance was assessed using TOP-1 Accuracy (TOP-1 Acc) as a metric of classification accuracy, and computational efficiency was measured in terms of Floating Point Operations per Second (FLOPs) and the number of parameters (Params). All experiments were executed on NVIDIA Tesla A100 GPUs, each equipped with 80 GB of memory.

All experiments were conducted on NVIDIA Tesla A100 GPUs, each with 80 GB of memory.

\begin{figure}[t]
\centering
\includegraphics[width=0.95\columnwidth]{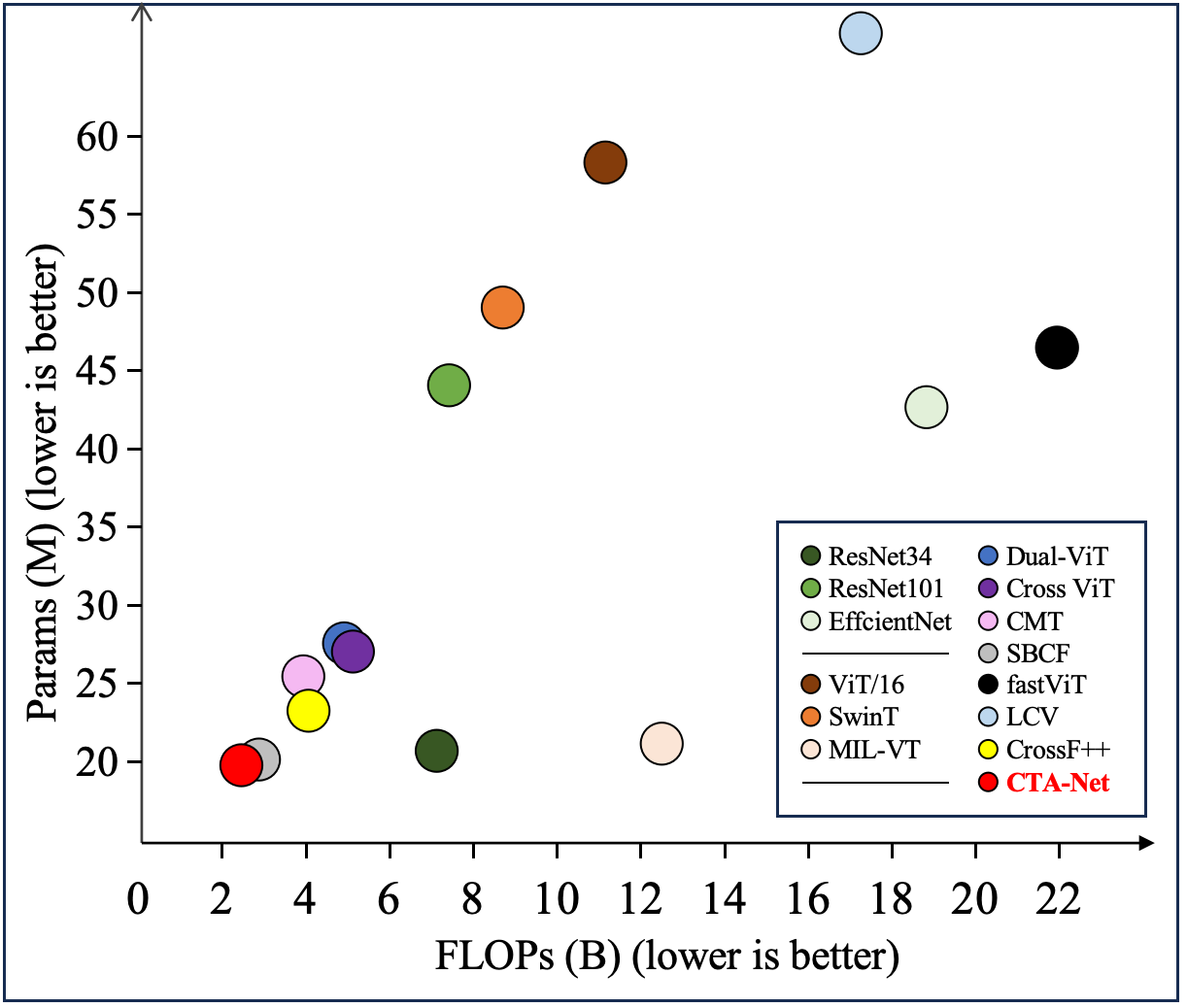}

\caption{Improvement of CTA-Net over CNN-Variants, ViT-Variants and ViT-Aggregation Model. Circles of different colors represent different models. The closer to the lower left corner, the smaller the model parameters and the higher the efficiency. \textbf{The red circle representing CTA-Net is closest to the lower left corner, and the model is the lightest and most efficient.}}
\label{fig3}
\end{figure}

\subsection{Comparison with State-Of-The-Art Methods}

Table \ref{table_1} presents the experimental results of CTA-Net on the four small-scale datasets. Compared to other CNN-Variants and ViT-Variants models, CTA-Net demonstrates superior performance. As illustrated in Figure \ref{fig3}, CTA-Net achieves excellent results with the lowest number of parameters and highest efficiency.  

\subsubsection{Comparisons with CNN-Variants Models.}

The experimental evaluation involved benchmarking CTA-Net against leading CNN and ViT models on four small-scale datasets, as detailed in Table \ref{table_1}. CTA-Net significantly outperforms several CNN-Variants. Notably, on the APTOS2019 and RFMiD2020 datasets, CTA-Net achieves 3.67\% and 5.1\% higher TOP-1 accuracy, respectively, than the average of three CNN-Variants. On the RFMiD2020 dataset, CTA-Net surpasses ResNet34 \cite{he:2016deep} by 9.22\%. These results underscore CTA-Net’s enhanced feature learning capabilities, with optimized parameter volume (20.32M) and FLOPs (2.83B), making it more efficient than traditional CNN architectures.

\subsubsection{Comparisons with ViT-Variants Models.}

As shown in Table \ref{table_1}, When compared to SOTA ViT models, CTA-Net exhibits remarkable performance, with an average increase in TOP-1 Acc of 12.07\%, 3.856\%, 21.52\%, and 12.93\% across the four datasets. On the CIFAR-10 and CIFAR-100 datasets, CTA-Net’s accuracy exceeds MIL-VT \cite{45:yu2021mil} by 37.76\% and 24.93\%, respectively, without the reliance on large-scale pre-trained weights. While CTA-Net's TOP-1 Acc on the CIFAR-100 dataset is slightly lower than SwinT’s, it compensates with significantly reduced FLOPs (5.87B less) and fewer parameters (29.24M less), achieving efficiency four times greater than SwinT \cite{24:swin-t}. These results emphasize CTA-Net’s balanced approach, leveraging CNN and ViT advantages to achieve high performance with fewer parameters and enhanced efficiency.

\subsubsection{Comparisons with ViT-Aggregation Models.}
Table \ref{table_1} presents a comparison between CTA-Net and various ViT-Aggregation models. CTA-Net’s superiority over ViT-Aggregation models is evident, with a 1.652\% average improvement in TOP-1 accuracy on small-scale datasets. It was found that fastViT converges very slowly. The TOP-1 Acc on the four small-scale datasets in Table 1 was achieved after training for 350 epochs, while other models were only trained for 100 epochs. The CTA-Net model converges faster and achieves higher performance within the same training cycles, even with limited data, showcasing its robust feature learning capabilities. Although Dual-ViT \cite{44:yao2023dual} slightly outperforms CTA-Net by 0.18\% in TOP-1 accuracy on CIFAR-10, CTA-Net demonstrates a 47.59\% higher efficiency and a 26.42\% reduction in parameters, which is crucial for resource-constrained environments. Similarly, while CrossF++/s \cite{31:crossformer++} achieves 90\% TOP-1 accuracy on CIFAR-10 with multi-epoch training, it requires significantly more computational resources, which conflicts with the practical need for balancing performance and efficiency. Additionally, It was observed that complex network structures like LCV \cite{13:ngo2024learning} encounter challenges on the small-scale CIFAR-10 dataset, achieving only 10\% TOP-1 accuracy without large-scale pre-trained weights (Not shown in Table 1). This points to the model's struggle in learning features from limited data.

In comparison to other aggregation models, CTA-Net not only delivers superior performance but also maintains the lowest parameter count (20.32M) and FLOPs (2.83B) among all baselines. This efficiency in feature learning and model deployment makes CTA-Net a compelling choice for applications involving small-scale datasets, improving Multi-Scale feature extraction and addressing the challenges of aggregation CNN and ViT architectures.

\subsection{Ablation Study}
To validate the effectiveness of CTA-Net, a series of ablation studies were conducted, focusing on the key innovative modules: the RRCV module and the LMF-MHSA module. The objective was to demonstrate how each component enhances the overall architecture's performance and identify the optimal configurations for integrating CNN and Transformer components.

\begin{table}[ht]
\centering

\begin{tabular}{ c c c c c }
\toprule[2pt]
{\multirow{2}*{\textbf{Model}}} & APTOS & RFMiD & CIFAR&CIFAR\\&2019&2020&10&100\\
\hline
\hline
ViT  &64.21\%  & 80.62\% &67.42\%& 41.41\% \\
+RRCV & 76.36\%&80.78\%  & 82.1\% & 55.73\%  \\
\textbf{+LMF-MHSA} & \textbf{76.9\%  }&\textbf{82.03\%}  &  \textbf{86.76\%}& \textbf{59.43\%}   \\
\toprule[2pt]

\end{tabular}
\caption{\textbf{Effectiveness of the Key Innovative Modules.} Metric is TOP-1 Acc. The first row indicates the use of a common vit model, the second row indicates the addition of our proposed RRCV module to this model, and the third row indicates the addition of our proposed LMF-MHSA module to the second row to form CTA-Net.}
\label{table_2}
\end{table}

\subsubsection{Effectiveness of the Key Innovative Modules.}
As depicted in Table \ref{table_2}, the RRCV and LMF-MHSA modules were incrementally added to the baseline to showcase their effectiveness. The addition of the RRCV module increased the TOP-1 Acc on the small-scale datasets by an average of 6.115\%, indicating that the RRCV module effectively integrates CNN's advantages and addresses ViT’s performance limitations on small-scale datasets. Further, incorporating the LMF-MHSA module resulted in an additional average TOP-1 Acc increase of 1.74\% across the four datasets, with a minimal increase in FLOPs from 2.48B to 2.83B. This showcases LMF-MHSA’s efficiency in handling multi-scale features.

\begin{table}[ht]
\centering

\begin{tabular}{ c c c c c}
\toprule[2pt]
{\multirow{2}*{\textbf{CNN-Variants}}} & APTOS & RFMiD & CIFAR&CIFAR\\&2019&2020&10&100\\

\hline
\hline
CNN  & 74.18\% & 80.62\% & 82.76\% &53.4\%\\

D-W Conv &76.09\%&81.25\% & 76.56\% & 49.6\% \\

\textbf{ResNet}  & \textbf{76.9\%} &\textbf{82.03\%}  &\textbf{86.76\%} & \textbf{59.43\%}\\

\toprule[2pt]
\end{tabular}
\caption{\textbf{Comparison on different CNN-Variants of RRCV Module}, Metric is TOP-1 Acc. D-W Conv refers to Depth-Wise Convolution. }
\label{table_3}
\end{table}

\subsubsection{Comparison on different CNN-Variants.}
The RRCV module embeds CNN operations within the Transformer architecture to enhance local feature extraction. Various configurations were tested, as shown in Table \ref{table_3}. Residual convolutions provided the best integration with the Transformer, maximizing performance, as detailed in Appendix B. This suggests that residual connections, which maintain gradient flow and support deeper models, are particularly beneficial for local feature extraction.


\begin{table}[ht]
\centering

\begin{tabular}{ c c c}
\toprule[2pt]
Module& FLOPs & Params \\

\hline
\hline
MHSA  & 11.28 B & 59.76 M  \\

\textbf{LMF-MHSA}  & \textbf{2.83 B} & \textbf{20.32 M}\\

\toprule[2pt]
\end{tabular}
\caption{Comparison between the LMF-MHSA module and the MHSA module in terms of model parameters and model efficiency.}
\label{table_4}
\end{table}

\subsubsection{Effectiveness of the Light Weight Multi-Scale Feature Fusion Multi-Head Self-Attention Module.}
The LMF-MHSA module was specifically designed to address parameter and computational efficiency. Table \ref{table_4} compares the traditional MHSA and LMF-MHSA under identical configurations. The LMF-MHSA module demonstrates a significant reduction in total parameter count to 20.83M, reducing the model complexity by 66\%. The model efficiency is increased to 2.83B, an increase of 79.42\%. showcasing its ability to maintain model performance while minimizing resource consumption. This efficiency highlights the module’s role in lightweight architecture design, facilitating its application in environments with limited computational capabilities.

\begin{table}[ht]
\centering

\begin{tabular}{ c c c c c}
\toprule[2pt]
{\multirow{2}*{\textbf{Module}}} & APTOS & RFMiD & CIFAR&CIFAR\\&2019&2020&10&100\\

\hline
\hline
1SSC & 74.46\% & 82.03\% & 84.2\%&58.84\%\\
3SSC &73.91\% & 81.09\% & 84.89\% &58.40\%\\
5SSC & 74.18\% & 81.25\% & 85.27\% &58.94\%\\
\textbf{MSC}  &\textbf{76.9\%  }&\textbf{82.03\%} & \textbf{86.76\%}&\textbf{59.43\%} \\

\toprule[2pt]

\end{tabular}
\caption{\textbf{Signle-Scale Conv vs. Multi-Scale Conv}, Metric is TOP-1 Acc. 1SSC refers to $1\times1$ Signle-Scale Conv. 3SSC refers to $3\times3$ Signle-Scale Conv. 5SSC refers to $5\times5$ Signle-Scale Conv. MSC refers to Multi-Scale Conv. }
\label{table_5}
\end{table}

\subsubsection{The Necessity of Multi-Scale Convolution.} The LMF-MHSA module employs multi-scale convolutions to significantly refine the feature extraction process. By enabling the network to capture information across varying levels of granularity, this approach is particularly effective for tasks requiring the recognition of intricate visual patterns. As demonstrated in Table \ref{table_5}, Experiments were conducted with different convolutional kernel sizes to validate the significance of multi-scale convolutions. Several attempts were made to experiment with single-scale convolution. For detailed experiments, refer to Appendix C. The results reveal that combining various kernel sizes in multi-scale convolutions yields an average performance improvement of 1.765\% over single-scale convolutions on small-scale datasets. This evidence underscores the critical role of multi-scale feature extraction in enhancing the model’s ability to generalize across diverse visual patterns. The integration of multiple convolutional kernels within the LMF-MHSA module facilitates a more robust feature representation, thereby boosting the overall performance of the CTA-Net architecture.

\section{Conclusion}

This paper presents CTA-Net, a CNN-Transformer aggregation network for improving multi-scale
feature extraction on small-scale datasets (fewer than 100,000). CTA-Net addresses the challenges of inadequate fusion of CNN and ViT features and high model complexity. By integrating CNN operations within the ViT framework, CTA-Net leverages the strengths of both architectures to enhance local feature extraction and global information processing, improving the network’s representation capabilities. The Reverse Reconstruction CNN Variant (RRCV) and Lightweight Multi-Scale Feature Fusion Multi-Head Self-Attention (LMF-MHSA) modules were validated through extensive ablation experiments. The results demonstrate that CTA-Net achieves a superior TOP-1 Acc of 86.76\% over the baseline, with higher efficiency (FLOPs 2.83B) and lower complexity (Params 20.32M). CTA-Net is a suitable aggregation network for small-scale datasets (fewer than 100,000), advancing visual tasks and providing a scalable solution for future recognition research and applications.


\bibliography{aaai25} 

\end{document}



\begin{center}
    \textbf{\Huge Appendix}
    \vspace{1cm}
\end{center}

\renewcommand{\thesection}{\Alph{section}}
\renewcommand{\thesubsection}{\thesection.\arabic{subsection}}
\renewcommand{\appendixname}{APPENDIX}

The structure of this supplementary material can be summarized as follows. Section A provides more dataset details. Section B shows detailed experiments on different CNN-Variants in the Reverse Reconstruction CNN-Variants (RRCV) module. Section C Comparison of Multi-Scale and Single-Scale Convolution on Small-Scale Datasets (fewer than 100,000).

\newcommand{\appendixtitle}[1]{%
  \section*{\centering \uppercase{Appendix \thesection\\#1}}%
  \addcontentsline{toc}{section}{Appendix \thesection: #1}%
}
\section*{\centering{A. Dataset Detailed Information}}

The APTOS 2019 Blindness Detection (APTOS2019) dataset \cite{40:APTOS2019} comprises 3,661 retinal fundus images, curated for the purpose of diagnosing diabetic retinopathy (DR). The dataset includes cases of both non-proliferative diabetic retinopathy (NPDR) and proliferative diabetic retinopathy (PDR). For experimental evaluation, the dataset was partitioned into three subsets: 2,929 images for training, 366 images for validation, and 366 images for testing. The images are presented in RGB format and vary in resolution, with common dimensions including 2416×1736, 1050×1050, and 819×614 pixels. Notably, severe NPDR is diagnosed when the patient exhibits two or more hallmark features of severe NPDR, while PDR is categorized into high-risk and non-high-risk cases.

The 2020 Retinal Fundus Multi-Disease Image Dataset (RFMiD2020) \cite{41:rfmid} contains 3,200 retinal fundus images, annotated with binary labels to indicate the presence (1) or absence (0) of pathological conditions. The dataset was divided into training (1,920 images), validation (640 images), and testing (640 images) sets. Similar to APTOS2019, the images in RFMiD2020 exhibit a range of resolutions, including 2144$\times$1424, 4288$\times$2848, and 2048$\times$1536 pixels. Table \ref{apx_table_1} provides a detailed breakdown of the distribution of diabetic retinopathy levels within the APTOS2019 and RFMiD2020 datasets. All images were resized to 224$\times$224 pixels prior to input into the network, facilitating cross-validation by splitting the dataset into five equal parts.

The CIFAR-10 and CIFAR-100 datasets \cite{16:cifar} are established benchmarks widely utilized for evaluating image classification algorithms. Both datasets consist of 60,000 color images, originally sized at 32x32 pixels. CIFAR-10 is composed of 10 distinct classes, with each class containing 6,000 images, partitioned into 50,000 training images and 10,000 testing images. CIFAR-100 expands on this structure, offering 100 classes with 600 images per class, similarly split into 50,000 for training and 10,000 for testing. In the context of this study, the images were upsampled to a resolution of 224$\times$224 pixels to align with the input requirements of contemporary deep learning architectures, such as Vision Transformers (ViT) \cite{dosovitskiy2020image} and Convolutional Neural Networks (CNN) \cite{he:2016deep}. This upsampling step is instrumental in allowing the model to capture finer details. Consequently, this process ensures both compatibility and enhanced feature extraction by increasing the effective resolution of the input images.

\begin{table}[htbp]
\centering
\begin{tabular}{c c c c}
\toprule[2pt] 

\textbf{Dataset} & \textbf{Label} & \textbf{DR Level} & \textbf{Count} \\
\hline
\hline
\multirow{5}{*}{APTOS2019} & 0 & no DR & 1805 \\
 & 1 & mild NPDR & 370 \\
 & 2 & moderate NPDR & 999 \\
 & 3 & severe NPDR & 193 \\
 & 4 & PDR & 294 \\
\hline
\multirow{2}{*}{RFMiD2020} & 0 & no DR & 2568 \\
 & 1 & yes DR & 632 \\
\toprule[2pt] 
\end{tabular}

\caption{Dataset Distribution}
\label{apx_table_1}
\end{table}

\begin{table}[ht]
\centering

\begin{tabular}{c c c c c c}
\toprule[2pt]

{\multirow{2}*{\textbf{Variants}}}&
\textbf{Batch}&
{\multirow{2}*{\textbf{Depth}}}&
{\multirow{2}*{\textbf{Heads}}}&
\textbf{APTOS}&
\textbf{RFMiD}\\
 &\textbf{Size}& & & \textbf{2019}&\textbf{2020}\\

\hline
\hline

\multirow{4}*{CNN}&32&8&4&71.86\%&79.84\%\\
&32&8&8&73.5\%&80.31\%\\
&16&12&4&74.18\%&80.62\%\\
&16&16&4&73.37\%&80.62\%\\
\hline
\multirow{4}*{DW-Conv}&32&8&4&71.2\%&81.25\%\\
&32&8&8&76.09\%&80\%\\
&16&12&4&74.18\%&80.47\%\\
&16&16&4&74.46\%&81.25\%\\
\hline
\multirow{4}*{ResNet}&32&8&4&73.1\%&80.16\%\\
&32&8&8&\textbf{76.9\%}&\textbf{82.03\%}\\
&16&12&4&75\%&81.41\%\\
&8&16&4&74.18\%&79.47\%\\

\toprule[2pt]

\end{tabular}

\caption{\textbf{Impact of different CNN variants in APTOS2019 and RFMiD2020 datasets}, Metric is TOP-1 Acc. DW-Conv refers to Depth-Wise convolution structures. }
\label{apx_table_2}
\end{table}

\begin{table}[ht]
\centering

\begin{tabular}{c c c c c c}
\toprule[2pt]

{\multirow{2}*{\textbf{Variants}}}&
\textbf{Batch}&
{\multirow{2}*{\textbf{Depth}}}&
{\multirow{2}*{\textbf{Heads}}}&
\textbf{CIFAR}&
\textbf{CIFAR}\\
 &\textbf{Size}& & & \textbf{10}&\textbf{100}\\

\hline
\hline

\multirow{4}*{CNN}&32&8&4&84.13\%&53.26\%\\
&32&8&8&84.56\%&55.42\%\\
&16&12&4&84.27\%&55.17\%\\
&16&16&4&83.97\%&54.66\%\\
\hline
\multirow{2}*{DW-Conv}&32&8&4&78.25\%&49.11\%\\
&32&8&8&76.56\%&49.6\%\\
\hline
\multirow{4}*{ResNet}&32&8&4&85.17\%&59.19\%\\
&32&8&8&\textbf{86.76\%}&\textbf{59.43\%}\\
&16&12&4&86.22\%&58.87\%\\
&8&16&4&86.5\%&58.4\%\\

\toprule[2pt]

\end{tabular}

\caption{\textbf{Impact of different CNN-Variants in CIFAR-10 and CIFAR-100 datasets}, Metric is TOP-1 Acc. DW-Conv refers to Depth-Wise convolution structures. }
\label{apx_table_3}
\end{table}

\begin{figure}[t]
\centering
\includegraphics[width=0.95\columnwidth]{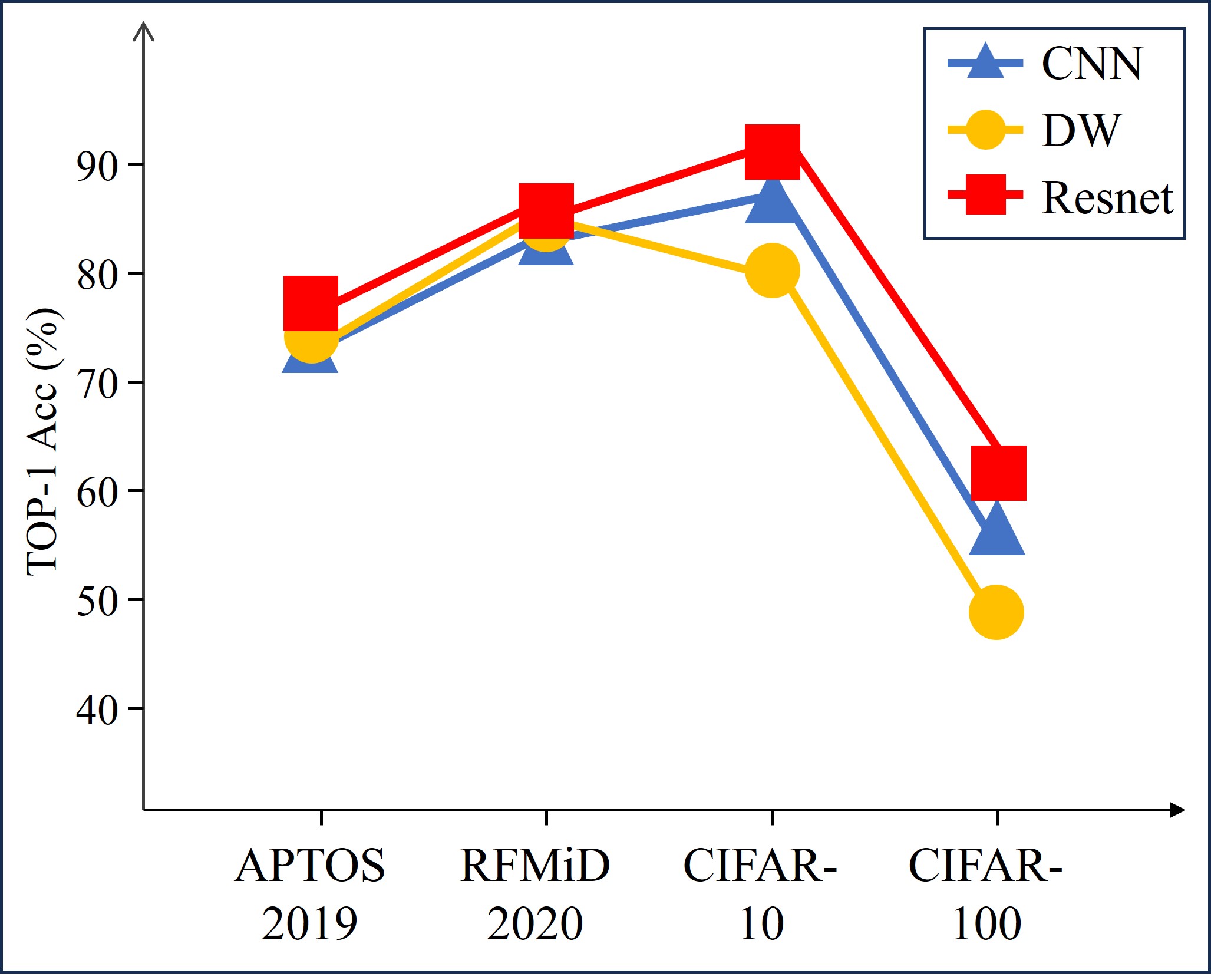}
\caption{Performance of different CNN-Variants.}
\label{apx_fig_1}
\end{figure}

\section*{\centering{B. Analysis of the Reverse Reconstruction CNN-Variants Module}}

The exploration of the Reverse Reconstruction CNN-Variants (RRCV) module is pivotal in assessing the influence of various CNN architectures on model performance, particularly within the constraints of small-scale datasets. This investigation seeks to pinpoint the most effective architecture for local feature extraction, a critical component in augmenting the representational capabilities of hybrid networks like CTA-Net.

The selection of different CNN variants, including standard CNN structures, Depth-Wise convolutional structures, and ResNet architectures \cite{he:2016deep}, is driven by their distinct architectural attributes and demonstrated efficacy in a range of computer vision tasks. Standard CNNs are employed as a baseline, providing a reference point for understanding conventional feature extraction mechanisms. Depth-Wise convolutions, recognized for their computational efficiency and reduced parameter count, offer insights into lightweight alternatives. ResNet architectures, with their hallmark skip connections, are designed to alleviate the vanishing gradient problem, thereby facilitating the training of deeper networks.

Experimental findings, as presented in Tables \ref{apx_table_2} and \ref{apx_table_3}, consistently demonstrate that the ResNet architecture outperforms other CNN variants. This superior performance is largely attributable to ResNet’s architectural innovations. The integration of identity mappings in ResNet facilitates efficient gradient flow and ensures the preservation of information across layers, which is crucial for maintaining the integrity of localized features within the RRCV module. Moreover, the deep hierarchical structure of ResNet allows for the development of more nuanced and rich feature representations, enhancing its capacity to extract local features while seamlessly integrating with the broader contextual information captured by the ViT components of CTA-Net.

Figure \ref{apx_fig_1} highlights the consistent performance gains observed when employing ResNet as a variant. On the APTOS2019 dataset, ResNet's TOP-1 accuracy surpasses that of standard CNNs by 1.57\% and Depth-Wise convolutional variants by 0.81\%. For the RFMiD2020 dataset, ResNet's TOP-1 accuracy exceeds that of standard CNNs by 0.42\% and Depth-Wise convolutions by 0.025\%. In the case of the CIFAR-10 dataset, ResNet achieves a 1.93\% higher TOP-1 accuracy compared to standard CNNs and an 8.67\% improvement over Depth-Wise convolutions. Similarly, on the CIFAR-100 dataset, ResNet outperforms standard CNNs by 4.35\% and Depth-Wise convolutions by 9.62\%. These results underscore ResNet's efficacy in enhancing feature extraction capabilities, establishing it as the optimal choice for the RRCV module within CTA-Net.

In conclusion, the empirical analysis emphasizes the criticality of architectural selection within the RRCV module, affirming that ResNet’s design is optimally suited for the extraction and preservation of local features. This design effectively complements the strengths of the ViT framework in CTA-Net, leading to an improved synergy that not only bolsters the network’s feature extraction prowess but also highlights the architectural advancements necessary for advancing visual recognition in resource-constrained environments.

\begin{table}[ht]
\centering

\begin{tabular}{c c c c c c}
\toprule[2pt]

\textbf{Conv}&
\textbf{Batch}&
{\multirow{2}*{\textbf{Depth}}}&
{\multirow{2}*{\textbf{Heads}}}&
\textbf{APTOS}&
\textbf{RFMiD}\\
 \textbf{Scale}&\textbf{Size}& & & \textbf{2019}&\textbf{2020}\\

\hline
\hline

\multirow{4}*{1$\times$1}&32&8&4&71.86\%&79.84\%\\
&32&8&8&74.46\%&82.03\%\\
&16&12&4&74.18\%&80.62\%\\
&16&16&4&73.37\%&81.23\%\\
\hline
\multirow{4}*{3$\times$3}&32&8&4&71.2\%&81.25\%\\
&32&8&8&73.91\%&81.09\%\\
&16&12&4&74.18\%&80.47\%\\
&16&16&4&73.46\%&81.25\%\\
\hline
\multirow{4}*{5$\times$5}&32&8&4&73.1\%&80.16\%\\
&32&8&8&74.18\%&81.25\%\\
&16&12&4&75\%&81.41\%\\
&8&16&4&74.18\%&79.47\%\\

\hline
\multirow{4}*{\textbf{MSC}}&32&8&4&74.23\%&81.66\%\\
&32&8&8&\textbf{76.9\%}&\textbf{82.03\%}\\
&16&12&4&75.87\%&82.02\%\\
&8&16&4&74.5\%&81.38\%\\

\toprule[2pt]

\end{tabular}

\caption{\textbf{Experimental results of single-scale convolution and multi-scale convolution under different parameters}, MSC refers to Multi-Scale Conv.}
\label{apx_table_4}
\end{table}

\begin{figure*}[t]
\centering
\includegraphics[width=0.90\linewidth]{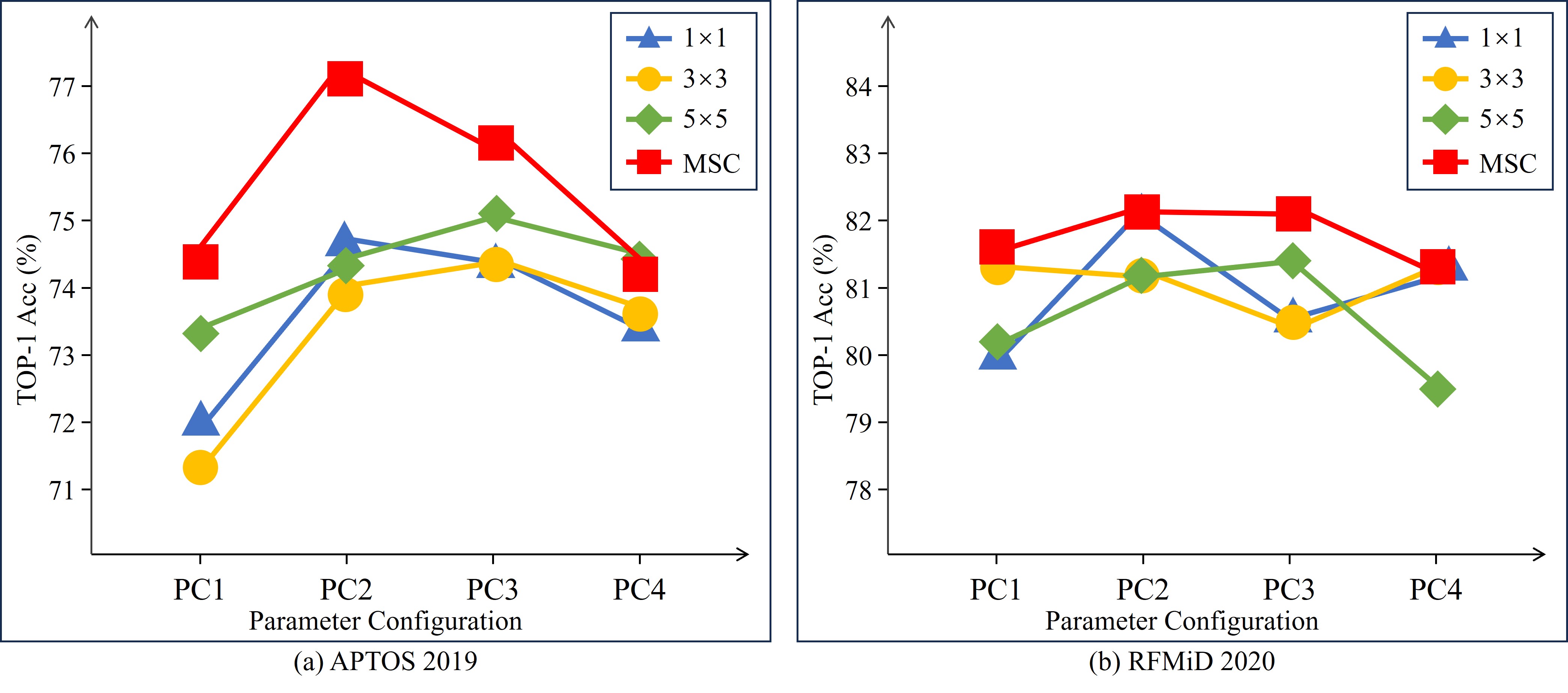}
\caption{\textbf{Comparison of the effects of multi-scale convolution and single-scale convolution on small-scale datasets.} (a) illustrates the comparative performance on the APTOS2019 dataset, while (b) depicts the results on the RFMiD2020 dataset. The horizontal axis in each figure denotes different parameter configurations. Specifically, PC1 refers to a configuration with a batch size of 32, a depth of 8, and 4 attention heads; PC2 corresponds to a batch size of 32, a depth of 8, and 8 attention heads; PC3 involves a batch size of 16, a depth of 12, and 4 attention heads; and PC4 denotes a setup with a batch size of 16, a depth of 16, and 4 attention heads.}
\label{apx_fig_2}
\end{figure*}

\section*{\centering{C. Single-Scale vs. Multi-Scale Convolution}}

The selection of convolutional kernel scale plays a crucial role in the performance of deep learning models, particularly in tasks requiring intricate feature extraction \cite{apx_0820_guo2020augfpn}. To explore this, a series of experiments using single-scale convolutional kernels of varying sizes (1x1, 3x3, and 5x5) were conducted on the APTOS2019 and RFMiD2020 datasets. The findings, as summarized in Table \ref{apx_table_4}, consistently demonstrate that models relying on any single kernel size underperform compared to those employing multi-scale convolutional operations.

This performance disparity can be attributed to the inherent limitations of single-scale convolutions. Single-scale operations restrict the model's ability to detect features only at a specific scale, potentially overlooking critical patterns that exist at other scales. For instance, finer details might be captured by smaller kernels, while larger patterns could be better recognized by larger kernels. However, a model restricted to a single kernel size lacks the flexibility to adapt to the diverse spatial complexities inherent in visual data \cite{apx_0820_ijgi13010005,apx_0820_guo2020augfpn}. In contrast, multi-scale convolutional operations effectively integrate kernels of different sizes, allowing the model to concurrently capture and synthesize features across multiple levels of granularity. This approach ensures a more comprehensive and robust feature representation, as the model can leverage the strengths of each kernel size to better interpret the data. The ability to capture diverse spatial information leads to significant improvements in the model's overall performance, as evidenced by the experimental results.

Moreover, as depicted in Figure \ref{apx_fig_2}, multi-scale convolution consistently yields superior outcomes across various hyperparameter settings. The experiments reveal that within the same convolutional scale, optimal results are generally obtained with a batch size of 32, a depth of 8, and 8 attention heads. However, the most pronounced performance gains are observed when multi-scale convolution is employed, highlighting its critical role in enhancing the model's feature extraction capabilities.
In summary, the integration of multi-scale convolutional operations significantly enhances the model's ability to capture complex patterns within the data, making it a superior choice over single-scale convolutional methods for tasks requiring detailed and comprehensive feature extraction.



\bibliography{aaai25} 